\documentclass[sigconf]{acmart}

\usepackage[utf8]{inputenc}

\setcopyright{none}
\renewcommand\footnotetextcopyrightpermission[1]{}
\setcopyright{none}
\makeatletter
\renewcommand\@formatdoi[1]{\ignorespaces}
\makeatother
\AtBeginDocument{%
  }

\usepackage{enumitem}
\usepackage{footmisc}
\usepackage{threeparttable} 

\begin{document}

\title{Federated Learning with Graph-Based Aggregation for Traffic Forecasting} 

\author{Audri Banik}
\email{abanik@brocku.ca}
\affiliation{%
  \institution{Brock University}
  \city{St Catharines}
  \state{ON}
  \country{Canada}
}

\author{Glaucio Haroldo Silva de Carvalho}
\email{gdecarvalho@brocku.ca}
\affiliation{%
  \institution{Brock University}
  \city{St Catharines}
  \state{ON}
  \country{Canada}
}

\author{Renata Dividino}
\email{rdividino@brocku.ca}
\affiliation{%
  \institution{Brock University}
  \city{St Catharines}
  \state{ON}
  \country{Canada}
}

\renewcommand{\shortauthors}{Banik et al.}

\begin{abstract}
In traffic prediction, the goal is to estimate traffic speed or flow in specific regions or road segments using historical data collected by devices deployed in each area. Each region or road segment can be viewed as an individual client that measures local traffic flow, making Federated Learning (FL) a suitable approach for collaboratively training models without sharing raw data. In centralized FL, a central server collects and aggregates model updates from multiple clients to build a shared model while preserving each client’s data privacy. Standard FL methods, such as Federated Averaging (FedAvg), assume that clients are independent, which can limit performance in traffic prediction tasks where spatial relationships between clients are important. Federated Graph Learning methods can capture these dependencies during server-side aggregation, but they often introduce significant computational overhead. In this paper, we propose a lightweight graph-aware FL approach that blends the simplicity of FedAvg with key ideas from graph learning. Rather than training full models, our method applies basic neighbourhood aggregation principles to guide parameter updates, weighting client models based on graph connectivity. This approach captures spatial relationships effectively while remaining computationally efficient. We evaluate our method on two benchmark traffic datasets, METR-LA and PEMS-BAY, and show that it achieves competitive performance compared to standard baselines and recent graph-based federated learning techniques. 
\end{abstract}



\maketitle

\section{Introduction}

Federated Learning (FL) allows multiple devices to collaboratively train machine learning models without sharing raw data. Instead, only model updates are transmitted, helping to preserve user privacy. In a centralized FL setup, a server manages the process by collecting local model parameters from clients, aggregating them, and then sending the updated global model back to each client.

Traffic prediction is a practical application of FL that naturally involves graph structures to represent spatial and temporal relationships among clients. In such systems, sensors or devices are placed on different road segments to monitor and predict local traffic patterns. Each sensor can make predictions using its own historical data, but incorporating information from nearby sensors often improves accuracy. A central server can aggregate these local model updates to generate more accurate, system-wide traffic forecasts. However, traditional FL algorithms such as FedAvg~\cite{mcmahan2017communication}, FedProx~\cite{li2020federated}, and FedRep~\cite{collins2021exploiting} tend to perform best when client data is independent and identically distributed (IID). This assumption does not always hold in practice.

\begin{figure*}[t!]
    \centering
    \includegraphics[scale=0.27]{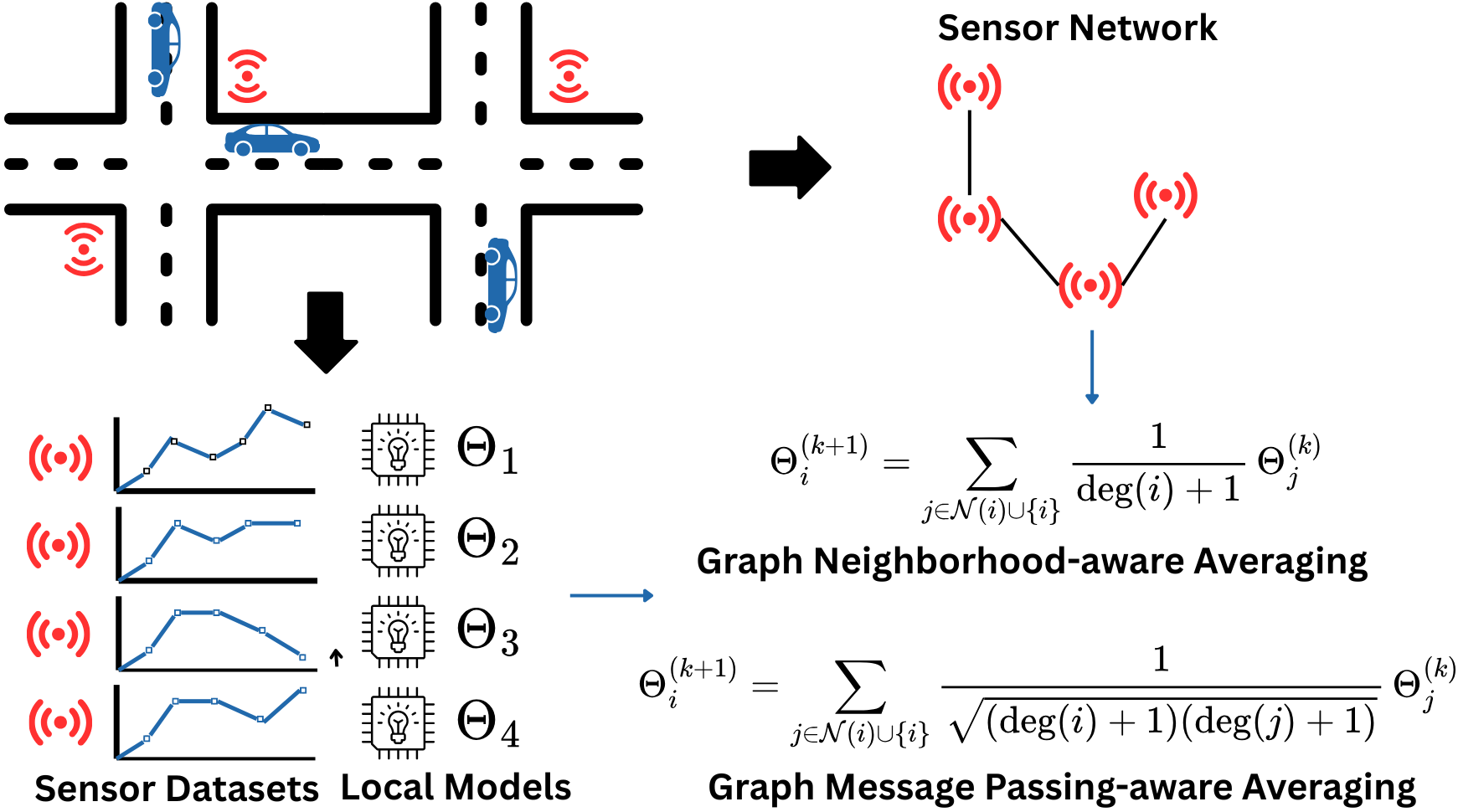}
  \caption{Illustration of a traffic forecasting system in a centralized FL system. Each road intersection is equipped with a traffic sensor running a local machine learning model to predict future traffic conditions. A central server periodically collects model parameters and performs aggregation using either Graph Neighbourhood-Aware Averaging (uniform averaging across connected intersections) or Graph Message Passing-Aware Averaging (degree-aware, structure-sensitive aggregation) on the graph of road connections.} 
  \label{fig:system}
\end{figure*}

Recent methods integrate graph models such as Graph Neural Networks (GNNs) into FL to better capture inter-client dependencies. These approaches use graphs to represent connectivity or similarity between clients, enabling more flexible aggregation strategies. In many cases, a global inter-client graph is maintained on the server, where it guides the training of a centralized model. The updated parameters are then sent back to the clients for the next training round. Recent approaches have demonstrated the effectiveness of using global graphs to model spatial and temporal relationships. These methods employ techniques such as graph-based representations, clustering, and attention mechanisms to improve model aggregation and support client personalization. For example, approaches like FedAGCN~\cite{10.1016/j.asoc.2023.110175} and FedASTA~\cite{li2024fedasta} build structured subgraphs or adaptive spatiotemporal graphs, enabling the server to guide learning based on evolving client relationships. FedDA~\cite{zhang2021dual} and FedGM~\cite{10633868} incorporate attention and meta-learning to refine aggregation and improve generalization across diverse regions. CNFGNN~\cite{c} demonstrates a collaborative modeling approach where the server handles spatial dependencies while clients focus on temporal patterns.  While these approaches improved overall performance by incorporating global information about client relationships, they rely on complex models and training pipelines, which can be computationally demanding and difficult to scale. 

To address these limitations, our work combines the simplicity of federated averaging with key ideas inspired by GNNs but avoids the complexity of full GNN training. Instead, we propose a lightweight graph-aware averaging method that captures essential relational patterns among clients using simple aggregation based on graph connectivity. Client model updates are weighted based on graph connectivity patterns, allowing federated averaging to incorporate client-to-client information flow across the graph. This approach offers a practical balance between performance and efficiency, making it suitable for resource-constrained environments.

We evaluated our approach on two benchmark traffic datasets: METR-LA and PEMS-BAY. Our experiments demonstrate that the proposed graph-based aggregation method is competitive and outperforms baseline approaches.

\section{Related Work}

In federated traffic prediction systems, the graph model can be located on the server or on the clients, resulting in two main system settings. 

\begin{description}[style=unboxed,leftmargin=0cm]
\item[Traffic Prediction Forecasting FL Systems: Server-Side Setting] 

In the server-side setting, a global inter-client graph representing relationships such as geographic proximity or shared characteristics is stored on the server and used to train a centralized graph-based model.  In FedAGCN~\cite{10.1016/j.asoc.2023.110175} the server uses a clustering algorithm to divide the transportation network into sub-graphs, where each sub-graph serves as a trainable client. The server then coordinates the GraphFed approach, which effectively utilizes the topological structure of transportation networks while the server manages computational time costs through asynchronous aggregation strategies. In the FedASTA framework~\cite{li2024fedasta}, while clients extract temporal and trend features from local time series data and upload them to the server, the server dynamically construct an adaptive spatial-temporal graph that captures evolving relationships among clients. The server applies a Filtered Fourier Transform and simplified Wasserstein-Fourier distance to measure similarity between nodes in the frequency domain. Using these server-computed distances, the server builds dynamic spatiotemporal graphs through nearest neighbor selection and constructs dynamic mask matrices for enhanced aggregation. In FedDA~\cite{zhang2021dual}, the server collects model updates from distributed clients and trains a quasi-global model that captures general temporal features. The server then organizes clients into clusters based on traffic pattern similarities and geographical locations. Within each cluster, the server implements an attention mechanism to selectively aggregate model updates, giving higher weights to clients with relevant information. The server applies a second attention mechanism to integrate knowledge across different clusters while maintaining sensitivity to their differences. The FedGM framework~\cite{10633868} the server employs the Reptile meta-learning algorithm for aggregation, processing local models from clients that consist of encoder-decoder modules and Graph Transformer Attention Networks. The server manages this complex aggregation while preserving privacy by ensuring clients avoid raw data sharing. In CNFGNN~\cite{c} the server takes responsibility for spatial modelling while clients handle temporal modelling. The server receives temporal embeddings from clients (processed using GRU-based encoders) and models spatial relationships using a two-layer graph neural network. The server then sends the resulting spatial embeddings back to clients for concatenation with local temporal embeddings. The server coordinates the alternating training between FedAvg for shared temporal feature extraction and Split Learning for server-side spatial model optimization.

The server-side setup allows the server to implement flexible aggregation strategies, aiming to balance local personalization with global generalization. However, training both global and local models can lead to convergence challenges and may become computationally expensive. Our work focuses on the server-side setting, proposing a lightweight averaging alternative approach over a global inter-client graph.

\item[Traffic Prediction Forecasting FL Systems: Client-Side Setting]

In the client-side setting, each client maintains its own local data (spatial and temporal) and and trains its local models using its own data. A common pattern emerges across client-side federated learning approaches for traffic prediction: the server consistently employs FedAvg or its variants for parameter aggregation, while the primary innovation occurs in client-side model architectures and the specific parameters being transmitted. The server aggregation strategy remains largely uniform across these approaches. In pFedCTP~\cite{zhang2024personalized}, the server applies standard FedAvg to temporal pattern representations. Similarly, the framework in~\cite{10155190} uses FedAvg for full model parameter aggregation. M$^3$FGM~\cite{tian2023m} employs FedAvg for model update aggregation, while~\cite{10137765} extends FedAvg with multilevel hierarchical processing. FedAGAT~\cite{ALHUTHAIFI2024120482} follows standard FedAvg procedures, and FedSTN~\cite{9737410} uses modified FedAvg adapted to edge computing environments. DSTGCN~\cite{10039323} implements FedAvg with opportunistic client selection, and communication optimization approaches like MFWA~\cite{articleFeng} and FedGCN-T~\cite{liu2024communication} apply weighted FedAvg and FedAvg to low-rank components, respectively. The key distinctions lie in client-side implementations and transmitted parameters. Clients in pFedCTP train personalized Spatio-Temporal Networks with hypernetwork-based adaptation, transmitting only temporal patterns rather than complete models. The framework in~\cite{10155190} has clients combine Graph Convolutional Networks with LSTM architectures, sending full model parameters. M$^3$FGM clients maintain dual-decoder architectures for collaborative and local predictions, while~\cite{10137765} employs Variational Graph Autoencoders for dynamic adjacency matrix generation. FedAGAT clients use Graph Attention Networks with spectral community detection, FedSTN clients train graph-based spatio-temporal models on edge devices, and DSTGCN clients implement Dynamic Spatio-Temporal Graph Convolutional Networks. Communication-optimized approaches vary in parameter representation: MFWA transmits performance-weighted parameters, while FedGCN-T sends only low-rank decomposed components. This pattern reveals that while most  traffic prediction focus heavily on client-side model design and parameter selection, the server-side aggregation remains anchored to FedAvg-based strategies, suggesting opportunities for server-side algorithmic enhancement.

Most recently, in~\cite{ijcai2024p611} clients employ Spatio-Temporal Neural Networks (ST-Net) that integrate GRU for temporal patterns and Graph Attention Networks for spatial dependencies. The server aggregates only the shared traffic pattern modules while clients keep spatial modules and predictors private. The server supports adaptive model interpolation by coordinating layer-wise strategies based on cosine similarity between global and local layers.  In FedSTG~\cite{10.1145/3701716.3715562}, clients train temporal models using GRU encoders, and the server implements FedATP, a personalized aggregation algorithm. The server processes complex graph structures extracted from client temporal evolution and spatial perspectives, operating two parallel Graph Convolutional Networks on static and evolutionary graphs. The server computes aggregation as a weighted sum based on similarity between client latent representations, with the server normalizing weights using softmax functions. While these work improve prediction, they introduce more complexity and training time.

\item[Graphs FL for Time-Series Tasks Beyond Traffic Forecasting] The PFGL framework \cite{li2024federated} aims to solve the forecasting problem in EV charging scenarios across geographically distributed stations. To model both spatial and behavioural relationships among clients, the authors construct two graphs on the server: a spatial graph that connects stations based on geographic closeness, and a behaviour one that connects clients whose local models behave similarly. Each client traint their time-series model locally which parameters are then sent to the server, where they are processed using a stacked Spatial-Temporal Graph Convolutional Network (ST-GCN) that aggregates spatial information from the Neighbourhood, and temporal patterns. Instead of relying on simple averaging (e.g., FedAvg), the server employs a weighted attention mechanism, where each client’s contribution is determined based on its model similarity and spatial closeness to others. In GFIoTL\cite{rasti2022graph}, the authors addresses smart home CIoT devices, aiming to tackle both statistical heterogeneity caused by non-IID data distributions and system heterogeneity. Each client trains a local model and periodically sends gradient updates to the server. The server applies a tunable graph filter (G-Fedfilt) implemented as a low-pass graph filter in the spectral domain, which is based on graph signal processing using the eigenvectors of the graph Laplacian, weighing contributions from structurally similar (neighboring) clients more than distant or dissimilar ones. FedBrain\cite{xie2024federated} works around differences in how hospitals split the brain into regions and how those regions link up to predict different diseases. On the client side, each hospital turns its fMRI/DTI scans into a brain connectivity graphs following graph construction schemes similar to BrainGB\cite{9933896}, where nodes represent Regions of Interest and edges reflect functional or structural connections derived from imaging data. On the server side, Fedbrain FedAvg sall the autoencoder weights into a global atlas mapper and pushes it back to every client. Additionally, hierarchical clustering is conducted to group hospitals with similar connectivity patterns.

\end{description}

\section{Preliminaries}
We consider a traffic forecasting scenario that naturally exhibits graph-structured data (see Figure~\ref{fig:system}). In this setting, a network of \( N \) traffic sensors is deployed across a road network to monitor and predict traffic conditions. Each client corresponds to a traffic sensor or a local region of sensors. The road network is modelled as a directed graph \( G = (V, E, W) \), where \( V \) is the set of sensor nodes with \( |V| = N \), and \( E \subseteq V \times V \) denotes the set of directed edges that capture road connectivity. The adjacency matrix \(A \in \{0,1\}^{N \times N}\) encodes the graph structure, where \(A_{ij} = 1\) indicates an edge between nodes \(v_i\) and \(v_j\), and \(A_{ij} = 0\) otherwise. Each client is associated with a \( d \)-dimensional feature vector. These vectors are stored in a feature matrix \( X \in \mathbb{R}^{N \times d} \), where the \( i \)-th row \( X_i \in \mathbb{R}^d \) corresponds to node \( v_i \).

Traffic observations collected from the sensor network are represented as a time series \( S \). At each time step \( t \), the signal \( S(t) \) captures the state of the network (e.g., traffic speed, volume, or occupancy). The goal of traffic forecasting is to learn a function \( h(\cdot) \) that, given the past \( T \) observations, predicts the next \( T \) future signals:
\[
[S(t - T + 1), \ldots, S(t)] \xrightarrow{h(\cdot)} [S(t + 1), \ldots, S(t + T)].
\]

Each client trains its own local time-series prediction model using only its local traffic data. The objective is to forecast future traffic conditions, ideally by accounting for both local and nearby traffic patterns. However, in practice, access to nearby data is often limited due to privacy constraints.

To address this, we explore how clients can benefit from spatial dependencies without directly accessing nearby traffic data. Rather than sharing raw observations with each other, clients periodically send model updates to a central server, which combines them using federated aggregation techniques. Through repeated cycles of local training and global aggregation, the clients work together to improve prediction accuracy while keeping their data private.

\section{Methodology}

We propose two graph-based FL aggregation methods that use the connectivity among clients to enhance model performance. Unlike traditional FL aggregation strategies like FedAvg~\cite{mcmahan2017communication}, which averages local model parameters uniformly, our approach accounts for spatial or relational dependencies between clients.

We model local client model parameters as node features over the client network, represented by a matrix \( X \in \mathbb{R}^{N \times P} \), where each row \( X_i \) contains the parameters of client \( i \). Our FL aggregation is guided by an adjacency matrix $A$ that captures the connectivity of the client or road network, with weights assigned based on topological relationships. Both graph-aware designs enable context-sensitive aggregation, enhancing accuracy while maintaining data locality and privacy. They provide a structured alternative to traditional averaging by leveraging local connectivity, making them well-suited for networked settings such as traffic sensors where client behaviour is influenced by neighbours. Multi-hop message passing further facilitates broader information exchange, leading to better performance and faster convergence.

\subsection{Graph Neighbourhood-Aware Averaging}

For our Graph Neighbourhood-Aware Averaging (GraphFedAvg), we extend the traditional FedAvg by incorporating the communication graph structure encoded in \( A \). To ensure that each client includes its own parameters during aggregation, we augment the adjacency matrix \( A \) with self-loops, \( \tilde{A} = A + I \), where \( I \) is the identity matrix. We define the degree matrix \( \tilde{D} \in \mathbb{R}^{N \times N} \) as a diagonal matrix where \( \tilde{D}_{ii} = \sum_{j} \tilde{A}_{ij} \).

Each client then aggregates the parameters of its neighbours (including itself) by computing a normalized weighted average. At each propagation step for \( \ell = 0, 1, \ldots, L-1 \), we update the client parameters as:
\[
X^{(\ell+1)} = \tilde{D}^{-1} \tilde{A} X^{(\ell)}
\]
where \( X^{(0)} \) represents the initial client parameters. Each application of \( \tilde{D}^{-1} \tilde{A} \) corresponds to one round of neighbourhood aggregation, and applying this operation \( L \) times enables information to propagate through \( L \)-hop neighbourhoods.

Equivalently, this can be expressed as replacing the traditional FedAvg global aggregation, which uses a simple average of local client models, with a weighted average over each client's neighbours. For each client \( i \in \{1, 2, \ldots, N\} \), the updated parameters are computed as:
\[
X_i^{(\ell+1)} = \sum_{j=1}^N \frac{\tilde{A}{ij}}{\tilde{D}{ii}} X_j^{(\ell)}
\]

\subsection{Graph Message Passing-Aware Averaging}
Inspired by the Label Propagation~\cite{zhur2002learning} algorithm commonly used in semi-supervised learning, our Graph Message Passing-Aware Averaging (MPFedAvg) method iteratively refines client parameters by blending their own values with those of their neighbours through the network leveraging the normalized adjacency matrix.
At each propagation step \(\ell = 1, \dots, L\), we update the parameters as given:
\[
X^{(\ell+1)} = \alpha \cdot \tilde{D}^{-1/2} \tilde{A} \tilde{D}^{-1/2} X^{(\ell)} + (1 - \alpha) X^{(\ell)}
\]
where \(\alpha \in [0, 1]\) is a hyperparameter controlling the trade-off between retaining the original local parameters and incorporating neighbourhood information.  In other words, this can be expressed as replacing the traditional aggregation simple averaging with a normalized average over each client's neighbours. For each client \( i \in {1, 2, \ldots, N} \), the updated parameters are computed as:

\[
X_i^{(\ell+1)} = \alpha \sum_{j=1}^N \frac{\tilde{A}_{ij}}{\sqrt{\tilde{D}_{ii} \mathstrut} \sqrt{\tilde{D}_{jj}}} X_j^{(\ell)} + (1 - \alpha) X_i^{(\ell)}
\]

This approach enables parameters propagation across the network, allowing clients to benefit from multi-hop neighbours while maintaining a weighted influence from their own initial models.

\section{Experiments}

\subsection{Datasets}
We demonstrate our model's effectiveness on traffic datasets, which exemplify spatio-temporal correlations and have been well-studied in forecasting research~\cite{a,c}. Our approach is generalizable to other spatio-temporal datasets with privacy concerns. We utilize two real-world datasets from~\cite{a}:
\begin{description}[style=unboxed,leftmargin=0cm]
    \item[PEMS-BAY:] Traffic speed readings from 325 sensors in the Bay Area collected over 6 months (Jan-May 2017).
    \item[METR-LA:] Traffic speed readings from 207 highway loop detectors in Los Angeles County over 4 months (Mar-Jun 2012).
\end{description}

For both datasets, the sensor adjacency matrix was constructed based on the road network distances using a thresholded Gaussian kernel (see~\cite{c} for more details). We aggregate readings into 5-minute windows and create sequences of length 24, with the forecasting task predicting the final 12 steps given the first 12.

\begin{center}
\begin{tabular}{lcccccc}
\toprule
\textbf{Dataset} & \textbf{Nodes} & \textbf{Edges} & \textbf{Train Seq} & \textbf{Val Seq} & \textbf{Test Seq} \\
\midrule
PEMS-BAY & 325 & 2,369 & 36,465 & 5,209 & 10,419 \\
METR-LA & 207 & 1,515 & 23,974 & 3,425 & 6,850 \\
\bottomrule
\end{tabular}
\end{center}

These methods are evaluated based on three commonly used metrics in traffic forecasting, including (1) Mean Absolute Error (MAE), (2) Mean Absolute Percentage Error (MAPE), and (3) Root Mean Squared Error (RMSE). Missing values are excluded in calculating these metrics. Detailed formulations of these metrics are provided in~\cite{c}.

\subsection{Clients Temporal Dynamics}

Each client employs a GRU-based encoder-decoder architecture~\cite{cho2014learning} to model node-level temporal dynamics. For input sequence $\mathbf{x}_i \in \mathbb{R}^{m \times D}$ at node $i$, the encoder processes the sequence to produce hidden state $\mathbf{h}_{c,i}$ as: $
\mathbf{h}_{c,i} = \text{Encoder}_i(\mathbf{x}_i, \mathbf{h}^{(0)}_{c,i})$, where $\mathbf{h}^{(0)}_{c,i}$ is initialized as a zero vector. The decoder then generates predictions $\hat{\mathbf{y}}_i$ auto-regressively, starting from the final input frame $x_{i,m}$ and utilizing the encoded context as: $\hat{\mathbf{y}}_i = \text{Decoder}_i(x_{i,m}, \mathbf{h}_{c,i})$. Training uses mean squared error (MSE) between predictions and ground truth values as the loss function, computed locally at each node.

\subsection{Server Aggregation Baselines}

We compare our proposed aggregation methods with the following baselines:

\begin{description}[style=unboxed,leftmargin=0cm]
    \item [GRU (centralized):] A GRU model trained on centralized sensor data. We use 2-layer GRU.
    \item [GRU (local):] A GRU model independently trained on each client's local data.
    \item [GRU + FedAvg:] A GRU model trained using the Federated Averaging algorithm \cite{mcmahan2017communication}.
    \item [GRU + FMTL:]  GRU-based encoder-decoder is used at the client side, and the server runs the federated multi-task learning (FMTL) with cluster regularization~\cite{smith2017federated} given by the adjacency matrix. 
    \item [CNFGNN:] A GRU-based encoder-decoder is used at the client side, while the server applies a 2-layer Graph Convolutional Network with residual connections to aggregate client models~\cite{c}.
\end{description}

All models are trained using the Adam optimizer with a learning rate of $1\mathrm{e}{-3}$, GRU has a hidden layer dimension $100$ and batch size of $128$. For the Graph Message Passing Aware (LPFedAvg) model we set $\alpha=0.8$. For experiments, we set local epochs to $3$, and client-server rounds to $5$.

\section{Results and Discussion}

\begin{table}[ht]
\small
\centering
\begin{threeparttable}
\caption{Comparison of performance on the traffic flow forecasting task. We use the Root Mean Squared Error (RMSE) to evaluate forecasting performance.}
\label{tab:forecasting_performance}
\begin{tabular}{lcc}
\toprule
\textbf{Method} & \textbf{PEMS-BAY} & \textbf{METR-LA} \\
\midrule
GRU (centralized, 2L)\tnote{*} & 4.172 & 11.787 \\
GRU (local, 2L)\tnote{*} & 4.152 & 12.224 \\
GRU (2L) + FedAvg\tnote{*} & 4.432 & 12.058 \\
GRU (2L) + FMTL\tnote{*} & 3.955 & 11.570 \\
CNFGNN \tnote{*}& 3.822 & 11.487 \\
\textbf{(Ours) GRU (2L) + GraphFedAvg (1L)} & 3.749 & 11.479  \\
\textbf{(Ours) GRU (2L) + GraphFedAvg (2L)} & 3.745 & \textbf{11.473} \\
\textbf{(Ours) GRU (2L) + MPFedAvg (1L)} & \textbf{3.733} & 11.489 \\
\textbf{(Ours) GRU (2L) + MPFedAvg (2L)} & 3.756 & 11.480 \\
\bottomrule
\end{tabular}
\begin{tablenotes}
\footnotesize
\item[*] Results as provided by~\cite{c}. We re-run the settings GRU (centralized), GRU (local), and GRU + FedAvg to confirm performance.
\end{tablenotes}
\end{threeparttable}
\end{table}

Table~\ref{tab:forecasting_performance} presents the RMSE results for traffic flow forecasting on the PEMS-BAY and METR-LA datasets. As expected, centralized training with GRU achieves strong performance (4.172 PEMS-BAY, 11.787 METR-LA), serving as an upper bound that demonstrates the potential when all data is available at a single location. Local training on individual clients shows weaker results (4.152 PEMS-BAY, 12.224 METR-LA) due to the absence of shared knowledge across clients, highlighting the fundamental challenge of federated learning in spatially-dependent domains.

FedAvg further degrades performance compared to local training on PEMS-BAY (4.432 vs 4.152), indicating that naive parameter averaging may not effectively capture the underlying relational structure among clients. This degradation occurs because FedAvg treats all clients equally during aggregation, ignoring the spatial relationships and varying traffic patterns across different geographical locations. 

Graph-aware approaches such as FMTL and CNFGNN show clear improvements over FedAvg by leveraging client similarity or topological information. FMTL (3.955 PEMS-BAY, 11.570 METR-LA) accounts for client heterogeneity, while CNFGNN (3.822 PEMS-BAY, 11.487 METR-LA) achieves lower RMSE than FMTL on both datasets, confirming the benefit of incorporating spatial modeling via server-side graph neural networks.

Our GraphFedAvg method introduces graph-aware parameter aggregation at the server level, where client updates are weighted based on their spatial proximity. The results demonstrate consistent improvements across both single-layer (3.749 PEMS-BAY, 11.479 METR-LA) and two-layer variants (3.745 PEMS-BAY, 11.473 METR-LA). The marginal difference between 1L and 2L variants (0.004 RMSE on PEMS-BAY, 0.006 on METR-LA) suggests that a single layer of graph convolution is sufficient to capture the essential spatial dependencies, making the 1L variant more computationally attractive. MPFedAvg extends the graph-aware concept by incorporating explicit message passing mechanisms during aggregation. The MPFedAvg results show the strongest performance with the 1L variant (3.733 PEMS-BAY, 11.489 METR-LA), achieving the lowest RMSE on PEMS-BAY. Interestingly, the 2L variant (3.756 PEMS-BAY, 11.480 METR-LA) shows slight degradation on PEMS-BAY but maintains competitive performance on METR-LA, suggesting that the optimal graph depth may be dataset-dependent. 

Our proposed methods outperform all baselines across both datasets, with improvements ranging from 1.9\% to 8.1\% over the next-best method (CNFGNN). The performance gains demonstrate that incorporating graph-aware averaging at the server can effectively preserve inter-client dependencies without incurring the computational overhead of training deep GNNs on the server.

\section{Conclusion}

In this work, we propose two lightweight, graph-aware aggregation strategies for federated traffic forecasting that capture spatial relationships between clients by leveraging neighborhood information during server-side model averaging, without adding significant computational cost. Our GraphFedAvg method introduces spatial proximity weighting through graph convolution operations, while MPFedAvg incorporates explicit message passing mechanisms to enable multi-hop information propagation between spatially-related clients. Both approaches retain the computational simplicity of FedAvg while improving performance through graph-based coordination that preserves inter-client dependencies.

We evaluate our methods on two benchmark traffic datasets, METR-LA and PEMS-BAY, demonstrating improvements of 1.9\% to 8.1\% over existing graph-based federated methods. The results show that our approaches outperform traditional FL baselines as well as more complex graph-based methods, with single-layer variants often matching or exceeding two-layer performance. This confirms their ability to model spatial dependencies efficiently while providing a scalable and computationally attractive alternative to more complex graph-based federated architectures.

As future work, we plan to extend our approach to dynamic graphs, where the relationships between clients change over time based on evolving traffic patterns and infrastructure modifications, which is common in real-world traffic systems.

\bibliographystyle{ACM-Reference-Format}
\bibliography{references}


\begin{thebibliography}{27}


\ifx \showCODEN    \undefined \def \showCODEN     #1{\unskip}     \fi
\ifx \showDOI      \undefined \def \showDOI       #1{#1}\fi
\ifx \showISBNx    \undefined \def \showISBNx     #1{\unskip}     \fi
\ifx \showISBNxiii \undefined \def \showISBNxiii  #1{\unskip}     \fi
\ifx \showISSN     \undefined \def \showISSN      #1{\unskip}     \fi
\ifx \showLCCN     \undefined \def \showLCCN      #1{\unskip}     \fi
\ifx \shownote     \undefined \def \shownote      #1{#1}          \fi
\ifx \showarticletitle \undefined \def \showarticletitle #1{#1}   \fi
\ifx \showURL      \undefined \def \showURL       {\relax}        \fi
\providecommand\bibfield[2]{#2}
\providecommand\bibinfo[2]{#2}
\providecommand\natexlab[1]{#1}
\providecommand\showeprint[2][]{arXiv:#2}

\bibitem[Al-Huthaifi et~al\mbox{.}(2024)]%
        {ALHUTHAIFI2024120482}
\bibfield{author}{\bibinfo{person}{Rasha Al-Huthaifi}, \bibinfo{person}{Tianrui
  Li}, \bibinfo{person}{Zaid Al-Huda}, {and} \bibinfo{person}{Chongshou Li}.}
  \bibinfo{year}{2024}\natexlab{}.
\newblock \showarticletitle{FedAGAT: Real-time traffic flow prediction based on
  federated community and adaptive graph attention network}.
\newblock \bibinfo{journal}{\emph{Information Sciences}}  \bibinfo{volume}{667}
  (\bibinfo{year}{2024}), \bibinfo{pages}{120482}.
\newblock
\showISSN{0020-0255}
\urldef\tempurl%
\url{https://doi.org/10.1016/j.ins.2024.120482}
\showDOI{\tempurl}


\bibitem[Cho et~al\mbox{.}(2014)]%
        {cho2014learning}
\bibfield{author}{\bibinfo{person}{Kyunghyun Cho}, \bibinfo{person}{Bart
  Van~Merri{\"e}nboer}, \bibinfo{person}{Caglar Gulcehre},
  \bibinfo{person}{Dzmitry Bahdanau}, \bibinfo{person}{Fethi Bougares},
  \bibinfo{person}{Holger Schwenk}, {and} \bibinfo{person}{Yoshua Bengio}.}
  \bibinfo{year}{2014}\natexlab{}.
\newblock \showarticletitle{Learning phrase representations using RNN
  encoder-decoder for statistical machine translation}.
\newblock \bibinfo{journal}{\emph{arXiv preprint arXiv:1406.1078}}
  (\bibinfo{year}{2014}).
\newblock


\bibitem[Collins et~al\mbox{.}(2021)]%
        {collins2021exploiting}
\bibfield{author}{\bibinfo{person}{Liam Collins}, \bibinfo{person}{Hamed
  Hassani}, \bibinfo{person}{Aryan Mokhtari}, {and} \bibinfo{person}{Sanjay
  Shakkottai}.} \bibinfo{year}{2021}\natexlab{}.
\newblock \showarticletitle{Exploiting shared representations for personalized
  federated learning}. In \bibinfo{booktitle}{\emph{International conference on
  machine learning}}. PMLR, \bibinfo{pages}{2089--2099}.
\newblock


\bibitem[Cui et~al\mbox{.}(2023)]%
        {9933896}
\bibfield{author}{\bibinfo{person}{Hejie Cui}, \bibinfo{person}{Wei Dai},
  \bibinfo{person}{Yanqiao Zhu}, \bibinfo{person}{Xuan Kan},
  \bibinfo{person}{Antonio Aodong~Chen Gu}, \bibinfo{person}{Joshua Lukemire},
  \bibinfo{person}{Liang Zhan}, \bibinfo{person}{Lifang He},
  \bibinfo{person}{Ying Guo}, {and} \bibinfo{person}{Carl Yang}.}
  \bibinfo{year}{2023}\natexlab{}.
\newblock \showarticletitle{BrainGB: A Benchmark for Brain Network Analysis
  With Graph Neural Networks}.
\newblock \bibinfo{journal}{\emph{IEEE Transactions on Medical Imaging}}
  \bibinfo{volume}{42}, \bibinfo{number}{2} (\bibinfo{year}{2023}),
  \bibinfo{pages}{493--506}.
\newblock
\urldef\tempurl%
\url{https://doi.org/10.1109/TMI.2022.3218745}
\showDOI{\tempurl}


\bibitem[Feng et~al\mbox{.}(2024a)]%
        {articleFeng}
\bibfield{author}{\bibinfo{person}{Jian Feng}, \bibinfo{person}{Cailing Du},
  {and} \bibinfo{person}{Qi Mu}.} \bibinfo{year}{2024}\natexlab{a}.
\newblock \showarticletitle{Traffic Flow Prediction Based on Federated Learning
  and Spatio-Temporal Graph Neural Networks}.
\newblock \bibinfo{journal}{\emph{ISPRS International Journal of
  Geo-Information}}  \bibinfo{volume}{13} (\bibinfo{date}{06}
  \bibinfo{year}{2024}), \bibinfo{pages}{210}.
\newblock
\urldef\tempurl%
\url{https://doi.org/10.3390/ijgi13060210}
\showDOI{\tempurl}


\bibitem[Feng et~al\mbox{.}(2024b)]%
        {10633868}
\bibfield{author}{\bibinfo{person}{Xinxin Feng}, \bibinfo{person}{Haoran Sun},
  \bibinfo{person}{Shunjian Liu}, \bibinfo{person}{Junxin Guo}, {and}
  \bibinfo{person}{Haifeng Zheng}.} \bibinfo{year}{2024}\natexlab{b}.
\newblock \showarticletitle{Federated Meta-Learning on Graph for Traffic Flow
  Prediction}.
\newblock \bibinfo{journal}{\emph{IEEE Transactions on Vehicular Technology}}
  \bibinfo{volume}{73}, \bibinfo{number}{12} (\bibinfo{year}{2024}),
  \bibinfo{pages}{19526--19538}.
\newblock
\urldef\tempurl%
\url{https://doi.org/10.1109/TVT.2024.3441759}
\showDOI{\tempurl}


\bibitem[Hu et~al\mbox{.}(2023)]%
        {10155190}
\bibfield{author}{\bibinfo{person}{Simon Hu}, \bibinfo{person}{Yin Ye},
  \bibinfo{person}{Qinru Hu}, \bibinfo{person}{Xin Liu},
  \bibinfo{person}{Shaosheng Cao}, \bibinfo{person}{Howard~H. Yang},
  \bibinfo{person}{Yongdong Shen}, \bibinfo{person}{Panagiotis Angeloudis},
  \bibinfo{person}{Leandro Parada}, {and} \bibinfo{person}{Chao Wu}.}
  \bibinfo{year}{2023}\natexlab{}.
\newblock \showarticletitle{A Federated Learning-Based Framework for
  Ride-Sourcing Traffic Demand Prediction}.
\newblock \bibinfo{journal}{\emph{IEEE Transactions on Vehicular Technology}}
  \bibinfo{volume}{72}, \bibinfo{number}{11} (\bibinfo{year}{2023}),
  \bibinfo{pages}{14002--14015}.
\newblock
\urldef\tempurl%
\url{https://doi.org/10.1109/TVT.2023.3287221}
\showDOI{\tempurl}


\bibitem[Li et~al\mbox{.}(2024b)]%
        {li2024fedasta}
\bibfield{author}{\bibinfo{person}{Kaiyuan Li}, \bibinfo{person}{Yihan Zhang},
  \bibinfo{person}{Huandong Wang}, \bibinfo{person}{Yan Zhuo}, {and}
  \bibinfo{person}{Xinlei Chen}.} \bibinfo{year}{2024}\natexlab{b}.
\newblock \showarticletitle{FedASTA: Federated adaptive spatial-temporal
  attention for traffic flow prediction}.
\newblock \bibinfo{journal}{\emph{arXiv preprint arXiv:2405.13090}}
  (\bibinfo{year}{2024}).
\newblock


\bibitem[Li et~al\mbox{.}(2020)]%
        {li2020federated}
\bibfield{author}{\bibinfo{person}{Tian Li}, \bibinfo{person}{Anit~Kumar Sahu},
  \bibinfo{person}{Manzil Zaheer}, \bibinfo{person}{Maziar Sanjabi},
  \bibinfo{person}{Ameet Talwalkar}, {and} \bibinfo{person}{Virginia Smith}.}
  \bibinfo{year}{2020}\natexlab{}.
\newblock \showarticletitle{Federated optimization in heterogeneous networks}.
\newblock \bibinfo{journal}{\emph{Proceedings of Machine learning and systems}}
   \bibinfo{volume}{2} (\bibinfo{year}{2020}), \bibinfo{pages}{429--450}.
\newblock


\bibitem[Li et~al\mbox{.}(2024a)]%
        {li2024federated}
\bibfield{author}{\bibinfo{person}{Yi Li}, \bibinfo{person}{Renyou Xie},
  \bibinfo{person}{Chaojie Li}, \bibinfo{person}{Yi Wang}, {and}
  \bibinfo{person}{Zhaoyang Dong}.} \bibinfo{year}{2024}\natexlab{a}.
\newblock \showarticletitle{Federated Graph Learning for EV Charging Demand
  Forecasting with Personalization Against Cyberattacks}.
\newblock \bibinfo{journal}{\emph{arXiv preprint arXiv:2405.00742}}
  (\bibinfo{year}{2024}).
\newblock


\bibitem[Li et~al\mbox{.}(2017)]%
        {a}
\bibfield{author}{\bibinfo{person}{Yaguang Li}, \bibinfo{person}{Rose Yu},
  \bibinfo{person}{Cyrus Shahabi}, {and} \bibinfo{person}{Yan Liu}.}
  \bibinfo{year}{2017}\natexlab{}.
\newblock \showarticletitle{Diffusion convolutional recurrent neural network:
  Data-driven traffic forecasting}.
\newblock \bibinfo{journal}{\emph{arXiv preprint arXiv:1707.01926}}
  (\bibinfo{year}{2017}).
\newblock


\bibitem[Liu et~al\mbox{.}(2023)]%
        {10137765}
\bibfield{author}{\bibinfo{person}{Lei Liu}, \bibinfo{person}{Yuxing Tian},
  \bibinfo{person}{Chinmay Chakraborty}, \bibinfo{person}{Jie Feng},
  \bibinfo{person}{Qingqi Pei}, \bibinfo{person}{Li Zhen}, {and}
  \bibinfo{person}{Keping Yu}.} \bibinfo{year}{2023}\natexlab{}.
\newblock \showarticletitle{Multilevel Federated Learning-Based Intelligent
  Traffic Flow Forecasting for Transportation Network Management}.
\newblock \bibinfo{journal}{\emph{IEEE Transactions on Network and Service
  Management}} \bibinfo{volume}{20}, \bibinfo{number}{2}
  (\bibinfo{year}{2023}), \bibinfo{pages}{1446--1458}.
\newblock
\urldef\tempurl%
\url{https://doi.org/10.1109/TNSM.2023.3280515}
\showDOI{\tempurl}


\bibitem[Liu et~al\mbox{.}(2024)]%
        {liu2024communication}
\bibfield{author}{\bibinfo{person}{Ruyue Liu}, \bibinfo{person}{Rong Yin},
  \bibinfo{person}{Xiangzhen Bo}, \bibinfo{person}{Xiaoshuai Hao},
  \bibinfo{person}{Xingrui Zhou}, \bibinfo{person}{Yong Liu},
  \bibinfo{person}{Can Ma}, {and} \bibinfo{person}{Weiping Wang}.}
  \bibinfo{year}{2024}\natexlab{}.
\newblock \showarticletitle{Communication-Efficient Personalized Federal Graph
  Learning via Low-Rank Decomposition}.
\newblock \bibinfo{journal}{\emph{arXiv preprint arXiv:2412.13442}}
  (\bibinfo{year}{2024}).
\newblock


\bibitem[McMahan et~al\mbox{.}(2017)]%
        {mcmahan2017communication}
\bibfield{author}{\bibinfo{person}{Brendan McMahan}, \bibinfo{person}{Eider
  Moore}, \bibinfo{person}{Daniel Ramage}, \bibinfo{person}{Seth Hampson},
  {and} \bibinfo{person}{Blaise~Aguera y Arcas}.}
  \bibinfo{year}{2017}\natexlab{}.
\newblock \showarticletitle{Communication-efficient learning of deep networks
  from decentralized data}. In \bibinfo{booktitle}{\emph{Artificial
  intelligence and statistics}}. PMLR, \bibinfo{pages}{1273--1282}.
\newblock


\bibitem[Meng et~al\mbox{.}(2021)]%
        {c}
\bibfield{author}{\bibinfo{person}{Chuizheng Meng}, \bibinfo{person}{Sirisha
  Rambhatla}, {and} \bibinfo{person}{Yan Liu}.}
  \bibinfo{year}{2021}\natexlab{}.
\newblock \showarticletitle{Cross-node federated graph neural network for
  spatio-temporal data modeling}. In \bibinfo{booktitle}{\emph{Proceedings of
  the 27th ACM SIGKDD conference on knowledge discovery \& data mining}}.
  \bibinfo{pages}{1202--1211}.
\newblock


\bibitem[Qi et~al\mbox{.}(2023)]%
        {10.1016/j.asoc.2023.110175}
\bibfield{author}{\bibinfo{person}{Tao Qi}, \bibinfo{person}{Lingqiang Chen},
  \bibinfo{person}{Guanghui Li}, \bibinfo{person}{Yijing Li}, {and}
  \bibinfo{person}{Chenshu Wang}.} \bibinfo{year}{2023}\natexlab{}.
\newblock \showarticletitle{FedAGCN: A traffic flow prediction framework based
  on federated learning and Asynchronous Graph Convolutional Network}.
\newblock \bibinfo{journal}{\emph{Appl. Soft Comput.}} \bibinfo{volume}{138},
  \bibinfo{number}{C} (\bibinfo{date}{May} \bibinfo{year}{2023}),
  \bibinfo{numpages}{11}~pages.
\newblock
\showISSN{1568-4946}
\urldef\tempurl%
\url{https://doi.org/10.1016/j.asoc.2023.110175}
\showDOI{\tempurl}


\bibitem[Rasti-Meymandi et~al\mbox{.}(2022)]%
        {rasti2022graph}
\bibfield{author}{\bibinfo{person}{Arash Rasti-Meymandi},
  \bibinfo{person}{Seyed~Mohammad Sheikholeslami}, \bibinfo{person}{Jamshid
  Abouei}, {and} \bibinfo{person}{Konstantinos~N Plataniotis}.}
  \bibinfo{year}{2022}\natexlab{}.
\newblock \showarticletitle{Graph federated learning for CIoT devices in smart
  home applications}.
\newblock \bibinfo{journal}{\emph{IEEE Internet of Things Journal}}
  \bibinfo{volume}{10}, \bibinfo{number}{8} (\bibinfo{year}{2022}),
  \bibinfo{pages}{7062--7079}.
\newblock


\bibitem[Smith et~al\mbox{.}(2017)]%
        {smith2017federated}
\bibfield{author}{\bibinfo{person}{Virginia Smith}, \bibinfo{person}{Chao-Kai
  Chiang}, \bibinfo{person}{Maziar Sanjabi}, {and} \bibinfo{person}{Ameet~S
  Talwalkar}.} \bibinfo{year}{2017}\natexlab{}.
\newblock \showarticletitle{Federated multi-task learning}.
\newblock \bibinfo{journal}{\emph{Advances in neural information processing
  systems}}  \bibinfo{volume}{30} (\bibinfo{year}{2017}).
\newblock


\bibitem[Tian et~al\mbox{.}(2023)]%
        {tian2023m}
\bibfield{author}{\bibinfo{person}{Yuxing Tian}, \bibinfo{person}{Jiachi Luo},
  \bibinfo{person}{Zheng Liu}, \bibinfo{person}{Song Li}, {and}
  \bibinfo{person}{Yanwen Qu}.} \bibinfo{year}{2023}\natexlab{}.
\newblock \showarticletitle{M 3 FGM: A Node Masking and Multi-granularity
  Message Passing-Based Federated Graph Model for Spatial-Temporal Data
  Prediction}. In \bibinfo{booktitle}{\emph{International Conference on Neural
  Information Processing}}. Springer, \bibinfo{pages}{551--566}.
\newblock


\bibitem[Wang et~al\mbox{.}(2022)]%
        {10039323}
\bibfield{author}{\bibinfo{person}{Hanqiu Wang}, \bibinfo{person}{Rongqing
  Zhang}, \bibinfo{person}{Xiang Cheng}, {and} \bibinfo{person}{Liuqing Yang}.}
  \bibinfo{year}{2022}\natexlab{}.
\newblock \showarticletitle{Federated Spatio-Temporal Traffic Flow Prediction
  Based on Graph Convolutional Network}. In \bibinfo{booktitle}{\emph{2022 14th
  International Conference on Wireless Communications and Signal Processing
  (WCSP)}}. \bibinfo{pages}{221--225}.
\newblock
\urldef\tempurl%
\url{https://doi.org/10.1109/WCSP55476.2022.10039323}
\showDOI{\tempurl}


\bibitem[Xie et~al\mbox{.}(2024)]%
        {xie2024federated}
\bibfield{author}{\bibinfo{person}{Han Xie}, \bibinfo{person}{Yi Yang},
  \bibinfo{person}{Hejie Cui}, {and} \bibinfo{person}{Carl Yang}.}
  \bibinfo{year}{2024}\natexlab{}.
\newblock \showarticletitle{Federated learning for cross-institution brain
  network analysis}. In \bibinfo{booktitle}{\emph{Medical Imaging 2024:
  Computer-Aided Diagnosis}}, Vol.~\bibinfo{volume}{12927}. SPIE,
  \bibinfo{pages}{106--119}.
\newblock


\bibitem[Yuan et~al\mbox{.}(2023)]%
        {9737410}
\bibfield{author}{\bibinfo{person}{Xiaoming Yuan}, \bibinfo{person}{Jiahui
  Chen}, \bibinfo{person}{Jiayu Yang}, \bibinfo{person}{Ning Zhang},
  \bibinfo{person}{Tingting Yang}, \bibinfo{person}{Tao Han}, {and}
  \bibinfo{person}{Amir Taherkordi}.} \bibinfo{year}{2023}\natexlab{}.
\newblock \showarticletitle{FedSTN: Graph Representation Driven Federated
  Learning for Edge Computing Enabled Urban Traffic Flow Prediction}.
\newblock \bibinfo{journal}{\emph{IEEE Transactions on Intelligent
  Transportation Systems}} \bibinfo{volume}{24}, \bibinfo{number}{8}
  (\bibinfo{year}{2023}), \bibinfo{pages}{8738--8748}.
\newblock
\urldef\tempurl%
\url{https://doi.org/10.1109/TITS.2022.3157056}
\showDOI{\tempurl}


\bibitem[Zhang et~al\mbox{.}(2021)]%
        {zhang2021dual}
\bibfield{author}{\bibinfo{person}{Chuanting Zhang}, \bibinfo{person}{Shuping
  Dang}, \bibinfo{person}{Basem Shihada}, {and} \bibinfo{person}{Mohamed-Slim
  Alouini}.} \bibinfo{year}{2021}\natexlab{}.
\newblock \showarticletitle{Dual attention-based federated learning for
  wireless traffic prediction}. In \bibinfo{booktitle}{\emph{IEEE INFOCOM
  2021-IEEE conference on computer communications}}. IEEE,
  \bibinfo{pages}{1--10}.
\newblock


\bibitem[Zhang et~al\mbox{.}(2024a)]%
        {zhang2024personalized}
\bibfield{author}{\bibinfo{person}{Yu Zhang}, \bibinfo{person}{Hua Lu},
  \bibinfo{person}{Ning Liu}, \bibinfo{person}{Yonghui Xu},
  \bibinfo{person}{Qingzhong Li}, {and} \bibinfo{person}{Lizhen Cui}.}
  \bibinfo{year}{2024}\natexlab{a}.
\newblock \showarticletitle{Personalized federated learning for cross-city
  traffic prediction}. In \bibinfo{booktitle}{\emph{33rd International Joint
  Conference on Artificial Intelligence, IJCAI}}. \bibinfo{pages}{5526--5534}.
\newblock


\bibitem[Zhang et~al\mbox{.}(2024b)]%
        {ijcai2024p611}
\bibfield{author}{\bibinfo{person}{Yu Zhang}, \bibinfo{person}{Hua Lu},
  \bibinfo{person}{Ning Liu}, \bibinfo{person}{Yonghui Xu},
  \bibinfo{person}{Qingzhong Li}, {and} \bibinfo{person}{Lizhen Cui}.}
  \bibinfo{year}{2024}\natexlab{b}.
\newblock \showarticletitle{Personalized Federated Learning for Cross-City
  Traffic Prediction}. In \bibinfo{booktitle}{\emph{Proceedings of the
  Thirty-Third International Joint Conference on Artificial Intelligence,
  {IJCAI-24}}}, \bibfield{editor}{\bibinfo{person}{Kate Larson}} (Ed.).
  \bibinfo{publisher}{International Joint Conferences on Artificial
  Intelligence Organization}, \bibinfo{pages}{5526--5534}.
\newblock
\urldef\tempurl%
\url{https://doi.org/10.24963/ijcai.2024/611}
\showDOI{\tempurl}
\newblock
\shownote{Main Track}.


\bibitem[Zhang et~al\mbox{.}(2025)]%
        {10.1145/3701716.3715562}
\bibfield{author}{\bibinfo{person}{Yudong Zhang}, \bibinfo{person}{Xu Wang},
  \bibinfo{person}{Xuan Yu}, \bibinfo{person}{Kuo Yang},
  \bibinfo{person}{Zhengyang Zhou}, {and} \bibinfo{person}{Yang Wang}.}
  \bibinfo{year}{2025}\natexlab{}.
\newblock \showarticletitle{FedSTG: Breaking through Spatio-Temporal Data Silos
  with Federated Graph Learning}. In \bibinfo{booktitle}{\emph{Companion
  Proceedings of the ACM on Web Conference 2025}} (Sydney NSW, Australia)
  \emph{(\bibinfo{series}{WWW '25})}. \bibinfo{publisher}{Association for
  Computing Machinery}, \bibinfo{address}{New York, NY, USA},
  \bibinfo{pages}{1534–1538}.
\newblock
\showISBNx{9798400713316}
\urldef\tempurl%
\url{https://doi.org/10.1145/3701716.3715562}
\showDOI{\tempurl}


\bibitem[Zhur and Ghahramanirh(2003)]%
        {zhur2002learning}
\bibfield{author}{\bibinfo{person}{Xiaojin Zhur} {and} \bibinfo{person}{Zoubin
  Ghahramanirh}.} \bibinfo{year}{2003}\natexlab{}.
\newblock \showarticletitle{Learning from labeled and unlabeled data with label
  propagation}.
\newblock  (\bibinfo{year}{2003}).
\newblock


\end{thebibliography}




\end{document}